\documentclass{article} % For LaTeX2e
\usepackage{iclr2018_conference,times}
\usepackage{hyperref}
\usepackage{url}
\usepackage{pdfpages}
\usepackage{amsfonts}
\usepackage{color}

\title{Compact Neural Networks based on the Multiscale Entanglement Renormalization Ansatz}
\author{Andrew Hallam$^1$, Edward Grant$^2$, Vid Stojevic$^{1,3}$, Simone Severini$^{2,4}$, Andrew G. Green$^{5}$ }

\iclrfinalcopy % Uncomment for camera-ready version, but NOT for submission.

\begin{document}
\maketitle
 \begin{center}\textit{$^1$ Department of Physics \& Astronomy, University College London \\ 
$^2$ Department of Computer Science, University College London \\ 
$^3$ GTN Ltd. \\
$^4$ Institute of Natural Sciences, Shanghai Jiao Tong University \\
$^5$ London Centre for Nanotechnology, University College London
}
\end{center}
\begin{abstract}
This paper demonstrates a method for tensorizing neural networks based upon an efficient way of approximating scale invariant quantum states, the Multi-scale Entanglement Renormalization Ansatz (MERA). We employ MERA as a replacement for the fully connected layers in a convolutional neural network and test this implementation on the CIFAR-10 and CIFAR-100 datasets. The proposed method outperforms factorization using tensor trains, providing greater compression for the same level of accuracy and greater accuracy for the same level of compression. We demonstrate MERA layers with $14000$ times fewer parameters and a reduction in accuracy of less than $1\%$ compared to the equivalent fully connected layers, scaling like $\mathcal{O} (N)$. 
 \end{abstract}

\section{Introduction}
%%%%%%%%%%%
The \emph{curse of dimensionality} is a major bottleneck in machine learning, stemming from the exponential growth of variables with the number of modes in a data set (\citet{2016arXiv160900893C}). Typically state-of-the-art convolutional neural networks have millions or billions of parameters. However, previous work has demonstrated that representations stored in the network parameters can be highly compressed without significant reduction in network performance (\citet{DBLP:journals/corr/NovikovPOV15}, \citet{DBLP:journals/corr/GaripovPNV16}, \citet{hinton2015distilling}). Determining the best network architecture for a given task remains an open problem.

Descriptions of quantum mechanical systems raise a similar challenge; representing $n$ $d$-dimensional particles requires a rank-$n$ tensor whose memory cost scales as $d^n$. Indeed, it was the promise of harnessing this that led Richard Feynman (\citet{Feynman1982-FEYSPW}) to suggest the possibility of quantum computation. In the absence of a quantum computer, however, one must use compressed representations of quantum states. 

A level of compression can be achieved by factorizing the tensorial description of the quantum wavefunction. Many such factorizations are possible, the optimal structure of the factorization being determined by the structure of correlations in the quantum system being studied. A revolution in quantum mechanics was made by realizing that the best way to characterize the distribution of correlations and information in a state is by a quantity known as \emph{entanglement} -- loosely the mutual quantum information between partitions of a quantum system (\citet{2010RvMP...82..277E}). 

This has led to many successful applications of tensorial approaches to problems in solid state physics and quantum chemistry over the past 25 years (\citet{2014AnPhy.349..117O}, \citet{2007arXiv0711.1398K}). Intriguing ideas have also emerged over the past few years attempting to bridge the successes of neural networks in machine learning with those of tensorial methods in quantum physics, both at a fundamental level (\citet{2017JSP...168.1223L}, \citet{2014arXiv1410.3831M}), and as a practical tool for network design (\citet{DBLP:journals/corr/LevineYCS17}). Recent work has  suggested that entanglement itself is a useful quantifier of the performance of neural networks (\citet{DBLP:journals/corr/LevineYCS17}, \citet{2017arXiv171004833L})

The simplest factorization employed in quantum systems is known as the \emph{matrix product state} (\citet{2014AnPhy.349..117O}). In essence, it expresses the locality of information in certain quantum states. It has already been adopted to replace expensive linear layers in neural networks -- in which context it has been independently termed \emph{tensor trains} (\citet{Oseledets:2011:TD:2079141.2079149}). This led to substantial compression of neural networks with only a small reduction in the accuracy (\citet{DBLP:journals/corr/NovikovPOV15}, \citet{DBLP:journals/corr/GaripovPNV16}). 

Here we use a different tensor factorization -- known as the \emph{Multi-scale Entanglement Renormalization Ansatz} (MERA) --  that encodes information in a hierarchical manner (\citet{2008PhRvL.101k0501V}). MERA works through a process of coarse graining or renormalization. There have been a number of papers looking at the relationship between renormalization and deep learning. MERA is a concrete realization of such a renormalization procedure (\citet{2009arXiv0912.1651V}) and so possesses a multi-scale structure that one might anticipate in complex data. A number of works have utilized tree tensor network models that possess a similar hierarchical structure. However, they do not include the \emph{disentangler} tensors that are essential if each layer of the MERA is to capture correlations on different length scales (\citet{2017arXiv171004833L}). 

In this work we employ MERA as a replacement for linear layers in a neural network used to classify the CIFAR-10 and CIFAR-100 datasets. Our results show that this performs better than the tensor train decomposition of the same linear layer, and gives better accuracy for the same level of compression and better compression for the same level of accuracy.   In Section 2 we introduce factorizations of fully connected linear layers, starting with the tensor train factorization followed by a tree-like factorization and finally the MERA factorization. In Section 3 we discuss how this is employed as a replacement for a fully connected linear layer in deep learning networks. Section 4 gives our main results and we note connections with the existing literature in Section 5. Finally, in Section 6 we discuss some potential developments of the work.

\section{Tensor Factorization of Linear Layers}
%%%%%%%%%%%%%%%%%%%%%
\begin{figure}[h]
\begin{center}
%\framebox[4.0in]{$\;$}
\includegraphics[width=0.6\textwidth]{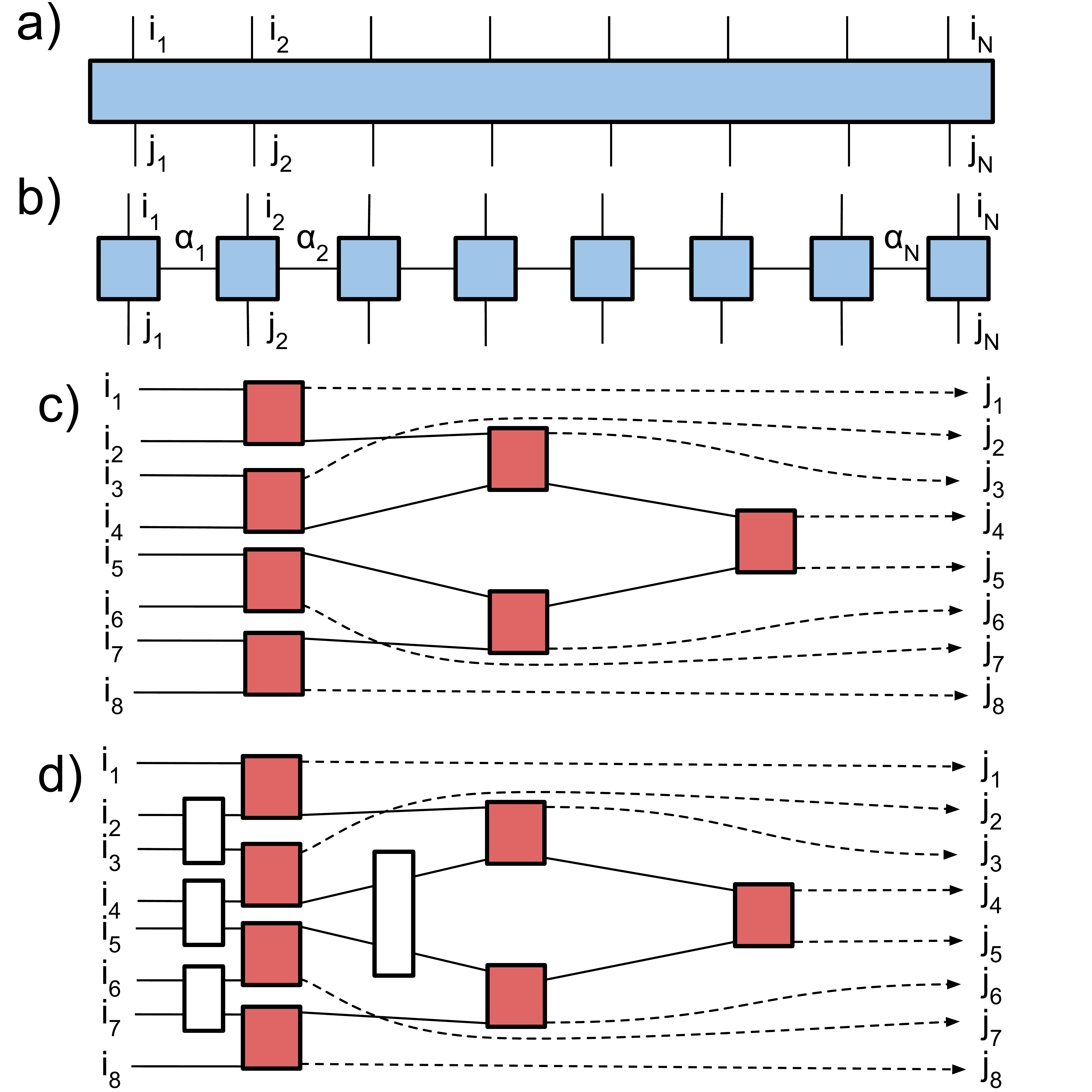}
\end{center}
\caption{Schematic diagrams of various tensor factorizations of linear layers. a) a general linear layer, b) its tensor train factorization. The squares represent smaller tensors. Connections represent contractions as indicated in Eq.(1). c) Tree network factorization. d) MERA factorization.}
\end{figure}
%%%%%%%%%%%%%%%%%%%%%

In this report we have replaced the linear layers of the standard neural network with tensorial MERA layers. The first step in achieving this involves expressing a linear layer as a tensor. This can be accomplished by taking a matrix ${\cal W}$ and reshaping it to be a higher dimensional array. For example, suppose ${\cal W}$ is $d^n$ by $d^n$ dimensional with components ${\cal W}_{AB}$. It can be transformed into a rank $2n$ tensor by mapping $A$ to $n$ elements $A \rightarrow i_1,i_2,...,i_n$ and $B$ to another $n$ elements $B \rightarrow j_1,j_2,...,j_n$. In this case each of the elements of the new tensor will be of size $d$.

Figure 1a gives a graphical representation of this rank $2n$ tensor 
${\cal W}^{i_1,i_2,...,i_n}_{j_1,j_2,...,j_n}$. It is important to note that in this representation, the lines represent the indices of the tensors rather than weights. Figure 1b illustrates the tensor train decomposition of ${\cal W}$. This consists of writing the larger tensor as the contraction of a \emph{train} of smaller tensors:
\begin{equation}
{\cal W}^{i_1,i_2,...,i_n}_{j_1,j_2,...,j_n}
= \sum_{\alpha_1,\alpha_2,...,\alpha_{n-1}}
A^{i_1}_{j_1, \alpha_1}
A^{\alpha_1,i_1}_{j_1, \alpha_2}
\;\; \cdot \cdot \cdot \;\;
A^{\alpha_{n-1},i_n}_{j_n}.
\label{tensortrain}
\end{equation}
In the tensor graphical notation, closed legs represent indices being summed over and free legs represent indices that aren't being summed over. For example, in equation 1 the $\alpha_i$ indices are being summed over and in Figure 1b the $\alpha_i$ lines are connected to tensors at both ends.

If each index runs over values from $1$ to $d$, this represents an exponential reduction from $d^{2n}$ parameters to $n(Dd)^2$, where the indices $\alpha$ run over values from $1$ to $D$ (known as the \emph{bond order} or \emph{Schmidt rank} in the quantum context). As noted above, this type of tensor factorization works well in physics when the information has a local structure (\citet{2010RvMP...82..277E}, \citet{2006PhRvB..73i4423V}); tensor trains capture correlations effectively up to length scales of order $\log D$ (\citet{2011AnPhy.326...96S}). This means that while useful for many tasks, the learned representations will be highly local. Tensors at either end of a tensor train decomposition of a linear layer will not be strongly correlated with one another. 

%Image data often possesses a more hierarchical structure. 

A hierarchically structured tensor network can better represent correlations across the linear layer. The tree tensor network shown in Figure 1c represents one possible hierarchical factorization. Each element of this network is a rank 4 tensor. The two tensors on the top left would have the form ${\cal M}^{j_1,\alpha_1}_{i_1,i_2}$ and ${\cal N}^{j_2,\alpha_2}_{i_3,i_4}$. The $i_n$ elements being represented by the lines on the left of the figure, the $j_n$ elements represented by the dotted lines on the right of the figure and the $\alpha_n$ lines being those connected with the tensor immediately to the right of ${\cal M}$ and ${\cal N}$.

Reading from left to right Figure 1c can be interpreted as follows: the tree-like connectivity imbues the network with a causal structure whereby a given linear element and its outputs are influenced by inputs in a region determined by its height in the tree.

For example, the rightmost element in Figure 1c is influenced by all of the inputs, whereas the top element in the middle column is influenced by inputs $i_1$ to $i_4$. Elements other than the rightmost tensor have one dashed output (that connects directly to the overall output) and one solid output (that ties it to the branching tree structure). These dashed lines are controlled by representations occurring on a particular scale in the data.

Notice that removing these dashed lines, the network has a true tree structure and represents a coarse graining or renormalization of the network. In this case, the linear elements are the \emph{isometries} of the original MERA's definition (\citet{2008PhRvL.101k0501V,2009arXiv0912.1651V}).

The simple tree network, which has been studied before in the context of neural networks, has a major deficiency. At each branching, it partitions the system in two, so that \emph{in extremis}, the correlations between neighbouring inputs  -- for example $i_4$ and $i_5$ in Figure 1c -- are only controlled by the element at the end of the network. Requiring the higher elements in the tree-structure to capture correlations between neighbouring inputs restricts their ability to describe the longer length scale correlations you would hope to capture by using a hierarchical structure.  

The  MERA (\citet{2009arXiv0912.1651V}) factorization was introduced in order to solve this problem. As can be seen in Figure 1d it adds an additional set of rank 4 tensors called {\it disentanglers}. The MERA is constructed by taking a tree network and placing one of these rank 4 tensors ${\cal D}^{\beta_1,\beta_2}_{\gamma_1,\gamma_2}$ such that its right-going legs $\beta_1$ and $\beta_2$ connect to two adjacent tensors of the tree network. For example, if we consider the top left-most disentangler in Figure 1d it has elements ${\cal D}^{\beta_1,\beta_2}_{i_2,i_3}$ and connects to the tree elements ${\cal M'}^{j_1,\alpha_1}_{i_1,\beta_1}$ and ${\cal N'}^{j_2,\alpha_2}_{\beta_2,i_4}$ with $\beta_1$ and $\beta_2$ then being summed over. 

The role of the disentanglers is to cause all correlations on the same length scale to be treated similarly. For example, correlations between any two neighbouring input indices $i_{n}$ and $i_{n+1}$ will be captured by either the first row of tree elements or the disentanglers. This allows the later elements in the network to work at capturing longer range correlations. 

In summary, a rank-$N$ MERA layer can be constructed in the following manner:

\begin{enumerate}
    \item Create a tree tensor layer. For example, an $N=2^\tau$ tree can be constructed from $2^{\tau-1}$ rank-4 tree tensors ${\cal M}^{\beta_1,\beta_2}_{\gamma_1,\gamma_2}$ in the first layer, followed by $2^{\tau-2}$ tree tensors in the second layer until after $\tau$ layers there is only a single tree tensor.
    
    \item A set of disentanglers are introduced. These are rank-4 tensors ${\cal D}^{\beta_1,\beta_2}_{\gamma_1,\gamma_2}$ which are placed such that every disentangler is contracted with two neighbouring tree tensors in an upcoming layer of the tree tensor.  
\end{enumerate}

\section{Experiments $\&$ network structure} 
%%%%%%%%%%%%%%

We have considered the performance of a neural network with the two penultimate fully connected layers of the model replaced with MERA layers, similar to the \citet{DBLP:journals/corr/NovikovPOV15} study of compression of fully connected layers using tensor trains. We have quantified the performance of the MERA layer through comparisons with two other classes of networks: fully connected layers with varying numbers of nodes and tensor train layers with varying internal dimension. The three types of network are otherwise identical. 

The networks consisted of three sets of two convolutional layers each followed by max pooling layers with $3\times3$ kernels and stride $2$. The convolutional kernels were $3\times3$. There were 64 channels in all of the convolutional layers except for the input, which had three channels, and the last convolutional layer, which had 256 channels. The final convolutional layer was followed by two more hidden layers, these were either fully connected, MERA layers or TT-layers depending upon the network.  The first of these layers was of size $4096 \times x $, the second is of size $x \times 64 $.  For the MERA and TT networks, these layers were $4096 \times 4096 $ and $4096 \times 64 $.

The final layer had $10$ or $100$ nodes corresponding to the image classes in CIFAR-10 and CIFAR-100. Leaky rectified linear units (LReLU) were used on all layers except the final layer, with $leak=0.2$~(\citet{maas2013rectifier}).

During training, nodes in the final convolutional layer and the two first fully connected layers were dropped with probability $0.5$. The penultimate convolutional layer nodes were dropped with probability $0.2$~(\citet{srivastava2014dropout}). Batch-normalization was used on all layers after dropout and max pooling~(\citet{ioffe2015batch}). We did not use bias units.

Gaussian weight initialization was employed in the fully connected models with standard deviation equal to $\frac{1}{\sqrt{n_{in}}}$, where $n_{in}$ was the number of inputs~(\citet{he2015delving}). 

In this report we considered networks with two varieties of fully-connected layers. The first of these networks had a $4096 \times 4096 $ fully connected layer followed by one which was $4096 \times 64 $; this network was used as a benchmark against which the other models could be compared. The second network instead had a $4096 \times n $ fully connected layer followed by a $n \times 64 $ layer where $n=5$ for the CIFAR-10 network and $n=10$ for the CIFAR-100 network. We trained these network to compare the MERA and tensor train layers to a fully connected model with a comparable number of parameters, in order to evaluate how detrimental naive compression is to accuracy.

A schematic of the two MERA layers can be found in Figure 2. The input to the first MERA layer was reshaped in to a rank-12 tensor with each index being dimension $2$, as described in Section 2. The MERA layer was then constructed from a set of rank-$4$ tensors using the method described in Section 2. 

The first MERA layer works as follows: It contains a column of 6 rank-$4$ tree elements, followed by 3 tree elements and finally a single tree element. 5 disentanglers are placed before the first column of tree elements and 2 more disentanglers are placed before the second column of tree elements. 

The second MERA layer has an identical structure to the first MERA layer, one of the outputs of the first set of tree elements is fixed. As a result the output of the second MERA layer is $64$ nodes. 

MERA weights were initialized using elements of randomized orthogonal matrices (\citet{2013arXiv1312.6120S}). The tensors themselves were constructed by reshaping these matrices, as described in Section 2. The random orthogonal matrix was constructed using the method of Stewart (\citet{doi:10.1137/0717034}, \citet{9eafeb2573aa4d7a9d3f0f17ec8c9af5}). Starting from a random $n-1 \times n-1$ dimensional orthogonal matrix, a random $n \times n$ dimensional orthogonal matrix can be constructed by taking a randomly distributed $n$-dimensional vector, constructing its Householder transformation, and then applying the $n-1$ dimensional matrix to this vector.  

Finally, a network with its fully connected layers replaced with a tensor train decomposition was trained in order to provide a comparison with the MERA layers. The tensor train layers were constructed as described in Section 2 with the internal dimension being fixed at $D=3$. In the second tensor train layer, half of the output indices were fixed to match the second MERA layer.  

We tested performance on the CIFAR-10 and CIFAR-100 datasets. We used $45,000$ images for training, $5,000$ for validation and $10,000$ for testing. Each training batch consisted of $50$ images. Training data was augmented by randomly flipping and translating the input images by up to 4 pixels. Translated images were padded with zeros. All images were normalized by dividing by $255$ and subtracting the mean pixels value from the training set. 

Validation and test set accuracy was recorded every $500$ iterations and training was stopped when validation accuracy did not improve for $10$ successive tests. The network was trained using backpropagation and the Adam optimizer, with initial learning rate $0.001$~(\citet{kingma2014adam}) and a softmax-cross-entropy objective. The test set accuracy for the model with the highest validation set accuracy was recorded. Each network was trained $10$ times with a different random weight initialization.

The networks were implemented in Tensorflow r1.3 and trained on NVIDIA Titan Xp and 1080ti GPUs.

\begin{figure}
%[h]
\begin{center}
%%\framebox[4.0in]{$\;$}
\includegraphics[width=0.9\textwidth]{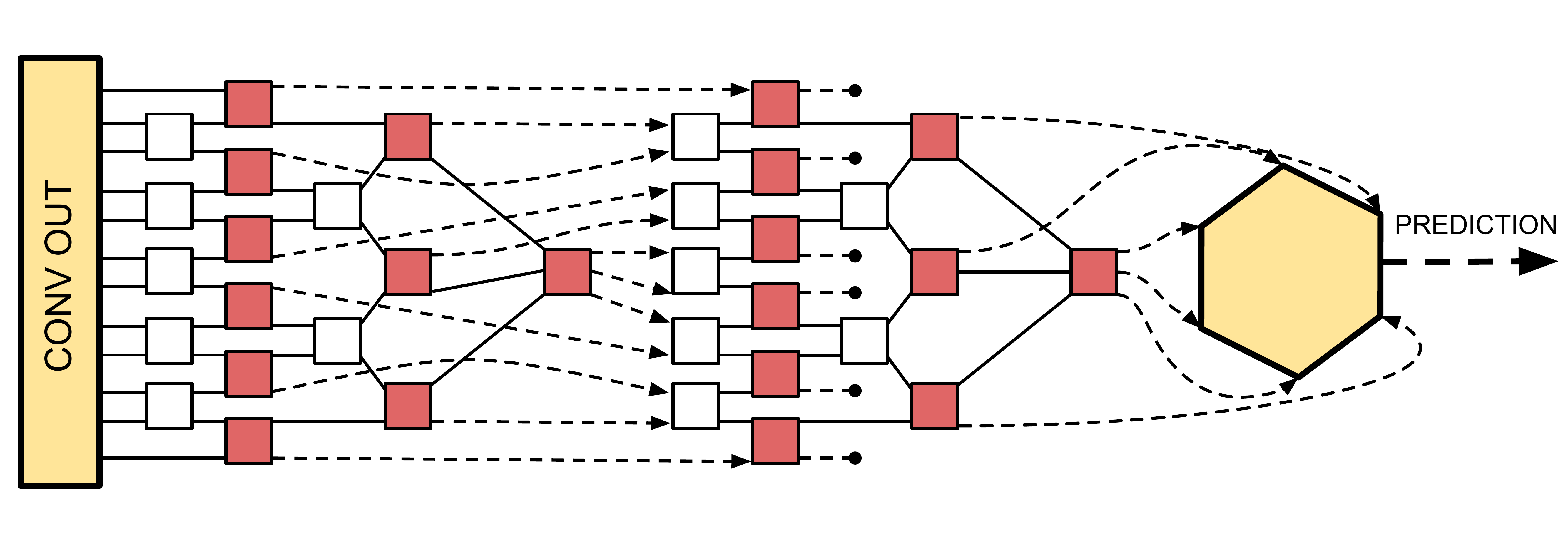}
\end{center}
\caption{A schematic of the MERA layers of the model. The small rectangles represent linear elements to factorize a general linear layer. White rectangles represent disentanglers. Red rectangles represent tree elements. Solid black lines connecting nodes represent tensor contraction and dashed lines with arrow heads represent the nonlinearities being applied. Dashed lines ending in a circle represent fixed outputs.}
\end{figure}
\section{Experimental results}
%%%%%%%%%%%%%%%
In Table 1 we compare the different models described in section 3 trained on the CIFAR-10 dataset. The compression rate stated is with respect to the number of parameters used in the fully-connected benchmark model, FC-1.

When comparing the MERA network to the fully connected model, FC-1 we see a considerable drop in the number of parameters required with only a modest drop in the accuracy of the network. MERA compresses the fully connected layers by a factor of $14,000$ with a drop in the accuracy of only $0.4\%$. We do not attempt to compress the convolutional layers in this work so in the MERA network the vast majority of the parameters are used in the convolutional layers which are identical to the fully connected model. 

How significant is the MERA network structure we have chosen to the results obtained? To test this we compare the MERA results obtained to the fully connected model with many fewer parameters in the fully connected layers, FC-2. Despite having around 20 times more parameters in the fully connected layer than the MERA model, the MERA model significantly out performs FC-2, with a $1.2\%$ drop in the accuracy of FC-2 compared to MERA. 

The MERA network also compares favourably to a tensor train network. In this case, the two networks have a comparable number of parameters but the MERA appears to achieve a higher accuracy than the tensor train network in this case. 

Results for the CIFAR-100 model can be seen in Table 2. While none of the networks are as accurate as the benchmark case, the MERA network continues to outperform the tensor train and ablated fully connected network. However, the reduction in accuracy compared to the fully connected network is larger than for the CIFAR-10 dataset.

In addition to the degree of compression achieved by these networks, we also address the time to optimize. There is evidently a degree of compromise required here: the number of multiplications required to apply a MERA layer scales with the input size $N$ and bond order $D$ as $N^{\log_2 D}$. The equivalent scaling for a tensor train and fully connected layer are $ND^2$ and $N^2$, respectively. This is reflected in the times taken to optimize these networks. Note however, that MERA can accommodate correlations at all scales of its input even at low bond order, whereas tensor trains require a bond order that scales exponentially with the length scale of correlation (\citet{2014AnPhy.349..117O}). MERA is, therefore, expected to scale better for very large data sets than either tensor trains or fully connected layers.

\begin{table}[h]
\caption{The CIFAR-10 experimental results for the different models. FC1 was the fully-connected model and FC2 was the fully-connected model with severely reduced number of parameters in the fully-connected layers. MERA are the result for the MERA inspired network. Finally TT is the tensor train model with the internal dimension being 3.}
\centering
    \bigskip
 \begin{tabular}{|c | c c c c c c|} 
\hline 
\rule{0pt}{2ex}    
 Network & \vtop{\hbox{\strut Parameters}\hbox{\strut (FC Layer)}} & \vtop{\hbox{\strut Parameters}\hbox{\strut (Total)}} & \vtop{\hbox{\strut Compression}\hbox{\strut (FC layer)}} & \vtop{\hbox{\strut Compression}\hbox{\strut (Total)}} & Accuracy & \vtop{\hbox{\strut Standard}\hbox{\strut Deviation}} \\ [2ex] 
 \hline
\rule{0pt}{2.5ex}    
FC-1 & 17,040,000 & 17,336,640 & 1 & 1 & 88.9 & 0.2 \\ [1ex]
 FC-2 & 21,440 & 318,080 & 795 & 54.5 & 86.5 & 0.8 \\ [1ex]
 MERA & 1192 & 297,832 & 14,295 & 58.21 & 88.5 & 0.1 \\ [1ex]
 TT & 1312 & 297,952 & 12,987 & 58.19 & 87.9 & 0.2 \\ [1ex]
 \hline
 \end{tabular}
\end{table}

\begin{table}[h]
\caption{The CIFAR-100 experimental results for the different models. FC1 was the fully-connected model and FC2 was the fully-connected model with severely reduced number of parameters in the fully-connected layers. MERA are the result for the MERA inspired network. Finally TT is the tensor train model with the internal dimension being 3.}
    \bigskip
\centering

 \begin{tabular}{|c | c c c c c c|} 
\hline 
\rule{0pt}{2ex}    
 Network & \vtop{\hbox{\strut Parameters}\hbox{\strut (FC Layer)}} & \vtop{\hbox{\strut Parameters}\hbox{\strut (Total)}} & \vtop{\hbox{\strut Compression}\hbox{\strut (FC Layer)}} & \vtop{\hbox{\strut Compression}\hbox{\strut (Total)}} & Accuracy & \vtop{\hbox{\strut Standard}\hbox{\strut Deviation}} \\ [2ex] 
 \hline
\rule{0pt}{2.5ex}    
FC-1 & 17,045,760 & 17,342,400 & 1	 & 1 & 61.8 & 0.7 \\ [1ex]
 FC-2 & 48,000 & 344,640 & 355 & 50.3 & 53.4 & 0.6 \\ [1ex]
 MERA & 6952 & 303,592 & 2451 & 57.12 & 58.4  & 0.6\\ [1ex]
 TT & 7072 & 303,712 & 2410 & 57.10 & 57.9 & 0.6 \\ [1ex]
 \hline
 \end{tabular}
\end{table}

\section{Related Work}
%%%%%%%%%%%%%%%%%%%%%%
Given how memory intensive deep neural networks typically are, substantial effort has been made to reduce number of parameters these networks require without significantly reducing their accuracy. Some of these have taken a similar approach to the MERA network described above, using tensor decompositions of the fully connected layers. 

These include the tensor train models of \citet{DBLP:journals/corr/NovikovPOV15} and \citet{DBLP:journals/corr/GaripovPNV16}. Here we have found replacing a fully connected linear layer with a MERA factorization resulted in superior accuracy for a comparable number of parameters.     

More directly related to this MERA model are a number of tree tensor network models (\citet{2017arXiv171004833L}, \citet{DBLP:journals/corr/LevineYCS17}).  As Section 2 explained, tree tensor networks inconsistently capture correlations on the same length scale, this is the reason for the introduction of disentanglers. Tree tensors do not possess these and we expect them to struggle to capture long range correlations as effectively as MERA. 
 
A MERA works through a process of coarse graining or renormalization. There have been a number of other papers looking at the relationship between renormalization and deep learning. \citet{2017JSP...168.1223L} argue that the effectiveness of deep neural networks should be thought of in terms of renormalization and \citet{2014arXiv1410.3831M} demonstrate an exact mapping between the variational renormalization group and restricted Boltzmann machines. In this report we have taken a different approach: only the fully connected layers of the network were replaced with MERA layers.

\section{Discussion}
%%%%%%%%%%%%%%%%%%%%%%
We have shown that replacing the fully connected layers of a deep neural network with layers based upon the multi-scale entanglement renormalization ansatz can generate significant efficiency gains with only small reduction in accuracy. When applied to the CIFAR-10 data we found the fully connected layers can be replaced with MERA layers with $14,000$ times less parameters with a reduction in the accuracy of less than $1\%$. The model significantly outperformed compact fully connected layers with $70-100$ times as many parameters. Moreover, it outperformed a similar replacement of the fully connected layers with tensor trains, both in terms of accuracy for a given compression and compression for a given accuracy. While the MERA layer resulted in a larger accuracy drop in the CIFAR-100 case, it still outperformed a comparable tensor train network. 

An added advantage --- not explored here --- is that a factorized layer can potentially handle much larger input data sets, thus enabling entirely new types of computation. Correlations across these large inputs can be handled much more efficiently by MERA than by tensor trains. Moreover, a compressed network may provide a convenient way to avoid over-fitting of large data sets. The compression achieved by networks with these factorized layers comes at a cost. They can take longer to train than networks containing the large fully connected layers due to the number of tensor contractions required to apply the factorized layer.

Our results suggest several immediate directions for future inquiry. Firstly, there are some questions about how to improve the existing model. For example, before the MERA layer is used the input is reshaped into a rank-12 tensor. There isn't a well defined method for how to perform this reshaping optimally and some experimentation is necessary. The best way to initialize the MERA layers is also still an open question.

The results presented here are a promising first step for using MERA in a more fundamental way.  Since MERA can be viewed as a coarse graining procedure (as explained in Section 2), and image data is often well represented in a hierarchical manner, one possibility would be to simply train a two-dimensional MERA directly on an image dataset, with no reference to a neural network. In \citet{2016arXiv160505775M} a similar idea was explored  with matrix product states being trained directly on MNIST.  An alternative possibility would be the replacement of just the convolutional layers of the network with a two-dimensional MERA. Both of these approaches would be closer in spirit to the fundamental ideas about the relationships between quantum physics and machine learning proposed in  \citet{2017JSP...168.1223L} and \citet{2014arXiv1410.3831M}.

Additionally, there has been some work using entanglement measures to explore how correlations are distributed in deep neural networks, and then utilizing these in order to optimize the design of networks (\citet{2017arXiv171004833L}, \citet{DBLP:journals/corr/LevineYCS17}).  It would be intriguing to explore such ideas using MERA, for example by using the concrete MERA model explored in this paper, or one of the more ambitious possibilities mentioned above.

We end by noting two facts: any variational approximation to a quantum wavefunction can be used to construct a replacement for linear layers of a  network. There are many examples and each may have its sphere of useful application. Moreover, quantum computers of the type being developed currently by several groups are precisely described by a type of tensor network (a finite-depth circuit - and one that may very soon be too large to manipulate classically) and could be used as direct replacement for linear layers in a hybrid quantum/classical neural computation scheme.

\subsubsection*{Acknowledgments}

This work was supported by the Engineering and Physical Sciences Research Council [grant number EP/P510270/1]. The Titan Xp used for this research was donated by the NVIDIA Corporation. We would like to thank Miles Stoudenmire for many enlightening discussions.

\bibliography{iclr2018_conference}

\begin{thebibliography}{23}
\providecommand{\natexlab}[1]{#1}
\providecommand{\url}[1]{\texttt{#1}}
\expandafter\ifx\csname urlstyle\endcsname\relax
  \providecommand{\doi}[1]{doi: #1}\else
  \providecommand{\doi}{doi: \begingroup \urlstyle{rm}\Url}\fi

\bibitem[Hinton et~al.(2015)Hinton, Vinyals, and Dean]{hinton2015distilling}
Geoffrey Hinton, Oriol Vinyals, and Jeff Dean.
\newblock Distilling the knowledge in a neural network.
\newblock \emph{arXiv preprint arXiv:1503.02531}, 2015.

\bibitem[{Cichocki} et~al.(2016){Cichocki}, {Lee}, {Oseledets}, {Phan}, {Zhao},
  and {Mandic}]{2016arXiv160900893C}
A.~{Cichocki}, N.~{Lee}, I.~V. {Oseledets}, A.-H. {Phan}, Q.~{Zhao}, and
  D.~{Mandic}.
\newblock {Low-Rank Tensor Networks for Dimensionality Reduction and
  Large-Scale Optimization Problems: Perspectives and Challenges PART 1}.
\newblock \emph{ArXiv e-prints arXiv:1609.00893}, September 2016.

\bibitem[{Eisert} et~al.(2010){Eisert}, {Cramer}, and
  {Plenio}]{2010RvMP...82..277E}
J.~{Eisert}, M.~{Cramer}, and M.~B. {Plenio}.
\newblock {Colloquium: Area laws for the entanglement entropy}.
\newblock \emph{Reviews of Modern Physics}, 82:\penalty0 277--306, January
  2010.
\newblock \doi{10.1103/RevModPhys.82.277}.

\bibitem[Feynman(1982)]{Feynman1982-FEYSPW}
R.~P. Feynman.
\newblock Simulating physics with computers.
\newblock \emph{International Journal of Theoretical Physics}, 21\penalty0
  (6):\penalty0 467--488, 1982.

\bibitem[Garipov et~al.(2016)Garipov, Podoprikhin, Novikov, and
  Vetrov]{DBLP:journals/corr/GaripovPNV16}
Timur Garipov, Dmitry Podoprikhin, Alexander Novikov, and Dmitry~P. Vetrov.
\newblock Ultimate tensorization: compressing convolutional and {FC} layers
  alike.
\newblock \emph{CoRR}, abs/1611.03214, 2016.
\newblock URL \url{http://arxiv.org/abs/1611.03214}.

\bibitem[He et~al.(2015)He, Zhang, Ren, and Sun]{he2015delving}
Kaiming He, Xiangyu Zhang, Shaoqing Ren, and Jian Sun.
\newblock Delving deep into rectifiers: Surpassing human-level performance on
  imagenet classification.
\newblock In \emph{Proceedings of the IEEE international conference on computer
  vision}, pp.\  1026--1034, 2015.

\bibitem[Ioffe \& Szegedy(2015)Ioffe and Szegedy]{ioffe2015batch}
Sergey Ioffe and Christian Szegedy.
\newblock Batch normalization: Accelerating deep network training by reducing
  internal covariate shift.
\newblock In \emph{International Conference on Machine Learning}, pp.\
  448--456, 2015.

\bibitem[{Kin-Lic Chan} et~al.(2007){Kin-Lic Chan}, {Dorando}, {Ghosh},
  {Hachmann}, {Neuscamman}, {Wang}, and {Yanai}]{2007arXiv0711.1398K}
G.~{Kin-Lic Chan}, J.~J. {Dorando}, D.~{Ghosh}, J.~{Hachmann}, E.~{Neuscamman},
  H.~{Wang}, and T.~{Yanai}.
\newblock {An Introduction to the Density Matrix Renormalization Group Ansatz
  in Quantum Chemistry}.
\newblock \emph{ArXiv e-prints arXiv:0711.1398}, November 2007.

\bibitem[Kingma \& Ba(2014)Kingma and Ba]{kingma2014adam}
Diederik Kingma and Jimmy Ba.
\newblock Adam: A method for stochastic optimization.
\newblock \emph{arXiv preprint arXiv:1412.6980}, 2014.

\bibitem[Levine et~al.(2017)Levine, Yakira, Cohen, and
  Shashua]{DBLP:journals/corr/LevineYCS17}
Yoav Levine, David Yakira, Nadav Cohen, and Amnon Shashua.
\newblock Deep learning and quantum entanglement: Fundamental connections with
  implications to network design.
\newblock \emph{CoRR}, abs/1704.01552, 2017.
\newblock URL \url{http://arxiv.org/abs/1704.01552}.

\bibitem[{Lin} et~al.(2017){Lin}, {Tegmark}, and
  {Rolnick}]{2017JSP...168.1223L}
H.~W. {Lin}, M.~{Tegmark}, and D.~{Rolnick}.
\newblock {Why Does Deep and Cheap Learning Work So Well?}
\newblock \emph{Journal of Statistical Physics}, 168:\penalty0 1223--1247,
  September 2017.
\newblock \doi{10.1007/s10955-017-1836-5}.

\bibitem[{Liu} et~al.(2017){Liu}, {Ran}, {Wittek}, {Peng}, {Bl{\'a}zquez
  Garc{\'{\i}}a}, {Su}, and {Lewenstein}]{2017arXiv171004833L}
D.~{Liu}, S.-J. {Ran}, P.~{Wittek}, C.~{Peng}, R.~{Bl{\'a}zquez Garc{\'{\i}}a},
  G.~{Su}, and M.~{Lewenstein}.
\newblock {Machine Learning by Two-Dimensional Hierarchical Tensor Networks: A
  Quantum Information Theoretic Perspective on Deep Architectures}.
\newblock \emph{ArXiv e-prints arXiv:1710.04833}, October 2017.

\bibitem[Maas et~al.(2013)Maas, Hannun, and Ng]{maas2013rectifier}
Andrew~L Maas, Awni~Y Hannun, and Andrew~Y Ng.
\newblock Rectifier nonlinearities improve neural network acoustic models.
\newblock In \emph{Proc. ICML}, volume~30, 2013.

\bibitem[{Mehta} \& {Schwab}(2014){Mehta} and {Schwab}]{2014arXiv1410.3831M}
P.~{Mehta} and D.~J. {Schwab}.
\newblock {An exact mapping between the Variational Renormalization Group and
  Deep Learning}.
\newblock \emph{ArXiv e-prints arXiv:1410.3831}, October 2014.

\bibitem[Novikov et~al.(2015)Novikov, Podoprikhin, Osokin, and
  Vetrov]{DBLP:journals/corr/NovikovPOV15}
Alexander Novikov, Dmitry Podoprikhin, Anton Osokin, and Dmitry~P. Vetrov.
\newblock Tensorizing neural networks.
\newblock \emph{CoRR}, abs/1509.06569, 2015.
\newblock URL \url{http://arxiv.org/abs/1509.06569}.

\bibitem[{Or{\'u}s}(2014)]{2014AnPhy.349..117O}
R.~{Or{\'u}s}.
\newblock {A practical introduction to tensor networks: Matrix product states
  and projected entangled pair states}.
\newblock \emph{Annals of Physics}, 349:\penalty0 117--158, October 2014.
\newblock \doi{10.1016/j.aop.2014.06.013}.

\bibitem[Oseledets(2011)]{Oseledets:2011:TD:2079141.2079149}
I.~V. Oseledets.
\newblock Tensor-train decomposition.
\newblock \emph{SIAM J. Sci. Comput.}, 33\penalty0 (5):\penalty0 2295--2317,
  September 2011.
\newblock ISSN 1064-8275.
\newblock \doi{10.1137/090752286}.
\newblock URL \url{http://dx.doi.org/10.1137/090752286}.

\bibitem[{Saxe} et~al.(2013){Saxe}, {McClelland}, and
  {Ganguli}]{2013arXiv1312.6120S}
A.~M. {Saxe}, J.~L. {McClelland}, and S.~{Ganguli}.
\newblock {Exact solutions to the nonlinear dynamics of learning in deep linear
  neural networks}.
\newblock \emph{ArXiv e-prints arXiv:1312.6120}, December 2013.

\bibitem[{Schollw{\"o}ck}(2011)]{2011AnPhy.326...96S}
U.~{Schollw{\"o}ck}.
\newblock {The density-matrix renormalization group in the age of matrix
  product states}.
\newblock \emph{Annals of Physics}, 326:\penalty0 96--192, January 2011.
\newblock \doi{10.1016/j.aop.2010.09.012}.

\bibitem[Srivastava et~al.(2014)Srivastava, Hinton, Krizhevsky, Sutskever, and
  Salakhutdinov]{srivastava2014dropout}
Nitish Srivastava, Geoffrey~E Hinton, Alex Krizhevsky, Ilya Sutskever, and
  Ruslan Salakhutdinov.
\newblock Dropout: a simple way to prevent neural networks from overfitting.
\newblock \emph{Journal of machine learning research}, 15\penalty0
  (1):\penalty0 1929--1958, 2014.

\bibitem[{Stoudenmire} \& {Schwab}(2016){Stoudenmire} and
  {Schwab}]{2016arXiv160505775M}
Miles {Stoudenmire} and D.~J. {Schwab}.
\newblock {Supervised Learning with Quantum-Inspired Tensor Networks}.
\newblock \emph{ArXiv e-prints arXiv:1605.05775}, May 2016.

\bibitem[{Verstraete} \& {Cirac}(2006){Verstraete} and
  {Cirac}]{2006PhRvB..73i4423V}
F.~{Verstraete} and J.~I. {Cirac}.
\newblock {Matrix product states represent ground states faithfully}.
\newblock \emph{Physical Review B: Condensed Matter and Materials}, 73\penalty0 (9):\penalty0 094423, March 2006.
\newblock \doi{10.1103/PhysRevB.73.094423}.

\bibitem[{Vidal}(2008)]{2008PhRvL.101k0501V}
G.~{Vidal}.
\newblock {Class of Quantum Many-Body States That Can Be Efficiently
  Simulated}.
\newblock \emph{Physical Review Letters}, 101\penalty0 (11):\penalty0 110501,
  September 2008.
\newblock \doi{10.1103/PhysRevLett.101.110501}.

\bibitem[{Vidal}(2009)]{2009arXiv0912.1651V}
G.~{Vidal}.
\newblock {Entanglement Renormalization: an introduction}.
\newblock \emph{ArXiv e-prints arXiv:0912.1651}, December 2009.

\end{thebibliography}
\bibliographystyle{iclr2018_conference}

\end{document}